\documentclass{article}

\usepackage{arxiv}

\usepackage[utf8]{inputenc}
\usepackage[T1]{fontenc} 
\usepackage{hyperref}     
\usepackage{url}          
\usepackage{booktabs}      
\usepackage{amsfonts}     
\usepackage{nicefrac}      
\usepackage{microtype}     
\usepackage{multirow}
\usepackage{graphicx}
\usepackage{lipsum}
\usepackage{float}

\title{Shallow Self-Learning for Reject Inference in Credit Scoring}

\author{
  Nikita Kozodoi\thanks{E-mail: \textit{nikita.kozodoi@hu-berlin.de}} \\
  Humboldt University of Berlin | Kreditech \\
  Berlin, Germany \\
  \And
  Panagiotis Katsas \\
  Kreditech \\
  Hamburg, Germany\\
  \And
  Stefan Lessmann \\
  Humboldt University of Berlin \\
  Berlin, Germany \\
  \And
  Luis Moreira-Matias \\
  Kreditech \\
  Hamburg, Germany\\
  \And
  Konstantinos Papakonstantinou \\
  Kreditech  \\
  Hamburg, Germany \\
}

\begin{document}

\maketitle

\begin{abstract}
Credit scoring models support loan approval decisions in the financial services industry. Lenders train these models on data from previously granted credit applications, where the borrowers' repayment behavior has been observed. This approach creates sample bias. The scoring model (i.e., classifier) is trained on accepted cases only. Applying the resulting model to screen credit applications from the population of all borrowers degrades model performance. Reject inference comprises techniques to overcome sampling bias through assigning labels to rejected cases. The paper makes two contributions. First, we propose a self-learning framework for reject inference. The framework is geared toward real-world credit scoring requirements through considering distinct training regimes for iterative labeling and model training. Second, we introduce a new measure to assess the effectiveness of reject inference strategies. Our measure leverages domain knowledge to avoid artificial labeling of rejected cases during strategy evaluation. We demonstrate this approach to offer a robust and operational assessment of reject inference strategies. Experiments on a real-world credit scoring data set confirm the superiority of the adjusted self-learning framework over regular self-learning and previous reject inference strategies. We also find strong evidence in favor of the proposed evaluation measure assessing reject inference strategies more reliably, raising the performance of the eventual credit scoring model.     
\end{abstract}


\section{Introduction}

Financial institutions use supervised learning to guide lending decisions. The resulting credit scoring models, also called scorecards, predict the probability of default (PD) -- an applicant's willingness and ability to repay debt \cite{verbraken2014development}. Loan approval decisions are made based on whether the scorecard predicts an applicant to be a repaying borrower (\textit{good} risks) or likely defaulter (\textit{bad} risks).

Scoring models are trained on data of accepted applicants. Their repayment behavior has been observed, which provides the labels for supervised learning. Inevitably, the sample of accepted clients (accepts) differs from the overall population of credit applicants. Accepts have passed the screening of the lender's scorecard, whereas the population also includes clients how have been denied credit by that scorecard (rejects) as well as customers who have not applied for credit with the focal lender. As a result, scoring models suffer from sample bias. Training a learning algorithm only on data from accepts deteriorates the accuracy of PD predictions when the corresponding scorecard is out into production for screening incoming credit applications \cite{siddiqi2012credit}. 

Reject inference refers to techniques that remedy sampling bias through inferring labels for rejects. Previous research has suggested several labeling approaches including naive strategies (e.g., label all rejects as \textit{bad} risks) and model-based techniques \cite{siddiqi2012credit}. However, empirical evidence concerning the value of reject inference and the efficacy of individual labeling strategies is scarce. Several studies use incomplete data, which only contain accepted cases (e.g. \cite{banasik2010reject,crook2004does}), do not have a labeled unbiased sample with both accepts and rejects (e.g., \cite{bucker2013reject}) or use synthetic data (e.g., \cite{joanes1993reject}). In addition, the data sets employed in prior studies are usually low-dimensional (e.g., \cite{maldonado2010semi}), which is not representative of the real-world credit scoring data used today \cite{wang2018hybrid}. Previous work is also geared towards a small set of classifiers comprising linear models and  support vector machines (SVM) \cite{anderson2013modified,li2017reject,maldonado2010semi}. Yet, there is much evidence that other learning algorithms (e.g., tree-based ensembles) outperform these methods in credit scoring \cite{lessmann2015benchmarking,wang2012empirical}.

The contribution of this paper is two-fold. First, we introduce a novel self-learning framework for reject inference in credit scoring applications. Our framework includes two different probabilistic classifiers for the training and labeling stages. The training stage benefits from using a strong learner such as gradient boosting. However, we suggest using a shallow learner for the labeling stage and show that it achieves higher calibration with respect to the true PD \cite{niculescu2005obtaining}. As a result, we maximize the precision of our model on the extreme quantiles of its output and minimize the noise introduced on newly labeled rejects. 

Second, we introduce a novel measure (denoted as \textit{kickout}) to assess reject inference methods in a reliable and operational manner. Aiming at labeling rejects to raise scorecard performance, the acid test of a reject inference strategy involves comparing a scorecard without correction for sample bias to a scorecard that has undergone reject-inference based correction on data from an unbiased sample of clients including both accepts and rejects with actual labels for both groups of clients. Such a sample would represent the operating conditions of a scorecard and thus uncover the true merit of reject inference \cite{crook2004does}. However, obtaining such a sample is very costly as it requires a financial institution to lend money to a random sample of applicants including high-risk cases that would normally be denied credit. Drawing on domain knowledge, the proposed \textit{kickout} measure avoids dependence on the actual labels of rejects and, as we establish through empirical experimentation, assesses the merit of a reject inference method more accurately than previous evaluation approaches. The data set used in this paper includes an unbiased sample containing both accepts and rejects, giving us a unique opportunity to evaluate a scorecard in its operating conditions. 

The paper is organized as follows. Section 2 reviews related literature on reject inference.  Section 3 revisits the reject inference problem, presents our self-learning framework, and introduces the kickout measure. Section 4 describes our experimental setup and reports empirical results. Section 5 concludes the paper.

\section{Literature Review}

The credit scoring literature has suggested different model-based and model-free approaches to infer labels of rejected cases. Some model-free techniques rely on external information such as expert knowledge to manually label rejects \cite{montrichard2007reject}. Another approach is to label all rejected cases as \textit{bad} risks \cite{siddiqi2012credit}, assuming that the default ratio among the rejects is sufficiently high. One other strategy is to obtain labels by relying on external performance indicators such as credit bureau scores or an applicant's outcome on a previous loan \cite{ash2002best,siddiqi2012credit}.

Model-based reject inference techniques rely on a scoring model to infer labels for rejects. Table \ref{ri_methods} depicts corresponding techniques, where we sketch the labeling strategy used in a study together with the base classifier that was used for scorecard development. Table \ref{ri_methods} reveals that most reject inference techniques have been tested with linear models such as logistic and probit regression.

The literature distinguishes several approaches towards model-based reject inference such as augmentation, extrapolation, bivariate models and others \cite{li2017reject}. Extrapolation refers to a set of techniques that use the initial scoring model trained on the accepts to label the rejected cases. For instance, hard cutoff augmentation labels rejects by comparing their model-estimated PDs to a predefined threshold \cite{siddiqi2012credit}. Parceling introduces a random component, separating the rejected cases into segments based on the range (e.g., percentile) of PDs. Instead of assigning labels based on the individual scores of rejects, they are labeled randomly within the identified segments based on the expected default rate for each score range. A drawback of such techniques is their reliance on the performance of the initial scoring model when applied to rejects.

Augmentation (or re-weighting) is based on the fact that applicants with a certain distribution of features appear in the training data disproportionately due to a non-random sample selection \cite{crook2004does}. Re-weighting refers to the techniques that train an additional model that separates accepts and rejects and predicts the probability of acceptance. These probabilities are then used to compute sampling weights for a scoring model.

Some scholars suggest using a two-stage bivariate probit model or two-stage logistic regression to perform reject inference \cite{banasik2003sample}. A bivariate model incorporates the Heckman's correction to account for a sample bias within the model, estimating both acceptance and default probability. These models assume linear effects within the logistic or probit regression framework.

Empirical studies have shown little evidence that reject inference techniques described above improve scorecard's performance \cite{banasik2005credit,chen2001economic,crook2004does,verstraeten2005impact}. Recently suggested alternatives rely on semi-supervised learning. For example, Maldonado et al have shown that self-learning with SVM outperforms well-known reject inference techniques such as ignoring rejects or labeling all rejects as \textit{bad} risks \cite{maldonado2010semi}. Their work is continued by \cite{li2017reject}, who propose a semi-supervised SVM that uses a non-linear kernel to train a model.

We follow recent studies and cast the reject inference problem in a semi-supervised learning framework. Our approach to solve the problem is a variation of self-learning adapted to a credit scoring context by extending the work of \cite{maldonado2010semi}.

\begin{table}[H]\centering
	\caption{Model-Based Reject Inference Methods}
	\label{ri_methods}
	\begin{tabular}{@{\extracolsep{3pt}} lll}
		\\[-4ex]
		\hline
		\multicolumn{1}{c}{Reference} & \multicolumn{1}{c}{Inference technique} & \multicolumn{1}{c}{Base model} \\
		\hline
		Reichert et al (1983) \cite{reichert1983exam}      &  LDA-based                  &    LDA  \\
		Joanes (1993) \cite{joanes1993reject}      &  Reclassification           &    LR \\
		Hand et al (1993) \cite{hand1993can}           &  Ratio prediction          &    - \\
		Hand et al (1993) \cite{hand1993can}           &  Rebalancing model          &    - \\
		Feelders (1999) \cite{feelders2000credit}    &  Mixture modeling           &    LR, QDA \\
		Banasik et al (2003) \cite{banasik2003sample}     &  Augmentation               &    LR, Probit \\
		Smith et al (2004) \cite{smith2004bayesian}     &  Bayesian network           &    Bayesian \\
		Crook et al (2004) \cite{crook2004does}         &  Reweigthing                &    LR \\
		Verstraeten et al (2005) \cite{verstraeten2005impact} &  Augmentation               &    LR \\
		Banasik et al (2005) \cite{banasik2005credit}     &  Augmentation               &    LR \\
		Fogarty (2006) \cite{fogarty2006multiple}   &  Multiple imputation        &    LR  \\
		Montrichard (2007) \cite{montrichard2007reject} &  Fuzzy augmentation         &    LR  \\
		Banasik et al (2007) \cite{banasik2007reject}     &  Augmentation               &    LR, Probit \\
		Banasik et al(2007) \cite{banasik2007reject}    &  Bivariate probit           &    Probit \\
		Kim et al (2007) \cite{kim2007technology}       &  Bivariate probit           &    - \\
		Banasik et al (2010) \cite{banasik2010reject}   &  Augmentation               &    Survival \\
		Maldonado et al (2010) \cite{maldonado2010semi}     &  Self-training              &    SVM \\
		Maldonado et al (2010) \cite{maldonado2010semi}     &  Co-training                &    SVM \\
		Maldonado et al (2010) \cite{maldonado2010semi}     &  Semi-supervised SVM        &    SVM \\
		Chen et al (2012) \cite{chen2012bound}         &  Bound and collapse         &    Bayesian  \\
		B\"ucker et al (2012) \cite{bucker2013reject}     &  Reweighting                &    LR  \\
		Siddiqi (2012) \cite{siddiqi2012credit}    &  Define as bad              &    - \\
		Siddiqi (2012) \cite{siddiqi2012credit}    &  Soft cutoff augmentation   &    - \\
		Siddiqi (2012) \cite{siddiqi2012credit}    &  Hard cutoff augmentation   &    - \\
		Siddiqi (2012) \cite{siddiqi2012credit}    &  Parceling                 &    - \\
		Siddiqi (2012) \cite{siddiqi2012credit}    &  Nearest neighbors          &    - \\
		Anderson et al (2013) \cite{anderson2013modified}  &  Mixture modeling           &    LR  \\
		Li et al (2017) \cite{li2017reject}          &  Semi-supervised SVM        &    SVM \\
		\hline
	\end{tabular}
\end{table}

\section{Methodology}

\subsection{Self-Learning for Reject Inference}

In reject inference, we are given a set of $n$ examples $x_1, ..., x_n \in \mathbb{R}^k$, where $k$ is the number of features. Set $X$ consists of $l$ accepted clients $x^a_1, ..., x^a_l \in X^a$ with corresponding labels $y^a_1, ..., y^a_l \in \{\textnormal{\textit{good}},\textnormal{\textit{bad}}\}$ and $m$ rejected examples $x^r_1, ..., x^r_m \in X^r$, whose labels are unknown. To overcome sampling bias and eventually raise scorecard accuracy, reject inference aims at assigning labels $y^r_{1}, ..., y^r_{m}$ to the rejected examples, which allows using the combined data for training a scoring model.

Standard self-learning starts with training a classifier $f(x)$ on the labeled examples $x^a_1, ..., x^a_l$ and using it to predict the unlabeled examples $x^r_1, ..., x^r_m$. Next, the subset of unlabeled examples $X^* \subset X^r$ with the most confident predictions is selected such that $f(x^*_i \in X^*) > \alpha$ or $f(x^*_i \in X^*) < 1 - \alpha$, where $\alpha$ is a probability threshold corresponding to a specified percentile of $f(x^*_i \in X^r)$. The selected rejects are labeled in accordance with the classifier's predictions. Cases obtained within this process are removed from $X^r$ and appended to $X^a$ to form a new labeled sample $X_1^a$. Finally, the classifier is retrained on $X_1^a$ and used to score the remaining cases in $X^r$. The procedure is repeated until all cases from $X^r$ are assigned labels or until certain stopping criteria are fulfilled \cite{rosenberg2005semi}.

Self-learning assumes that labeled and unlabeled examples in $X$ follow the same distribution \cite{rosenberg2005semi}. In a credit scoring context, $X^a$ and $X^r$ come from two different distributions because the scoring model employed by the financial institution separates accepts and rejects based on their feature values. The difference in distributions has negative consequences for self-learning: since the initial model is trained on a sample that is not fully representative of the unlabeled data, predictions of this model for the rejects are less reliable. The error is propagated through the iterative self-learning framework, which deteriorates the performance of the final model due to the incorrectly assigned labels. 

In this section, we describe a novel shallow self-training framework for reject inference that is geared toward reducing the negative effects of sample bias. The proposed framework consists of three stages: filtering, labeling and model training. We summarize the algorithm steps in Algorithm 2.

Within the proposed framework, we suggest to filter and drop some rejected cases before assigning them with labels. The goal of the filtering stage is two-fold. Firstly, we strive to remove rejected cases that come from the most different part of distribution compared to the accepts. Removing these cases would reduce the risk of error propagation, since predictions of the model trained on the accepts become less reliable as the distribution of cases to be predicted becomes more different from the one observed on the training data. Secondly, we remove rejects that are most similar to the accepted cases. Labeling such cases would potentially provide little new information for a scorecard and might even harm performance due to introducing noise. Therefore, the filtering stage is aimed at removing the cases that could have a negative impact of the scorecard performance.

The filtering is performed with isolation forest, which is a novelty detection method that estimates the normality of a specific observation by computing the number of splits required to isolate it from the rest of the data \cite{liu2008isolation}. We train isolation forest on all accepts in $X^a$ and use it to evaluate the similarity of the rejects in $X^r$. Next, we remove rejects that are found to be the most and least similar to the accepts by dropping cases within the top $\beta_t$ and bottom $\beta_b$ percentiles of the similarity scores. Algorithm 1 describes the filtering stage.

After filtering, we use self-learning with distinct labeling and training regimes to perform reject inference. While the scoring model is based on a tree-based algorithm (gradient boosting), we propose using a weak learner for labeling rejects because of its ability to produce better-calibrated predictions \cite{niculescu2005obtaining}. In this paper, we rely on L1-regularized logistic regression (L1) to label rejects.

\begin{table}[H]\centering
\begin{tabular*}{\textwidth}{ll}
\hline
\multicolumn{2}{l}{\textbf{Algorithm 1.} Isolation Forest for Filtering Rejected Examples} \\
\hline
1. & Train an isolation forest classifier $g(x)$ using all data in $X^a$. \\
2. & Use $g(x)$ to evaluate similarity scores of all unlabeled examples in $X^r$. \\
3. & Select a subset $X^* \subset X^r$ such that $g(x^*_i \in X^*) \in \lbrack \beta_b, \beta_t \rbrack$, where $\beta_b$ and $\beta_t$ are values of pre-defined \\
   & percentiles of $g(x^r_j \in X^r)$, $j = 1, ..., m$. \\
4. & Remove examples $X^*$ from $X^r$.\\
\hline
\end{tabular*}
\end{table}

\begin{figure}[H]\centering
\label{score_dist}
\includegraphics[width = 0.8\textwidth, trim={0 0 0 0}, clip]{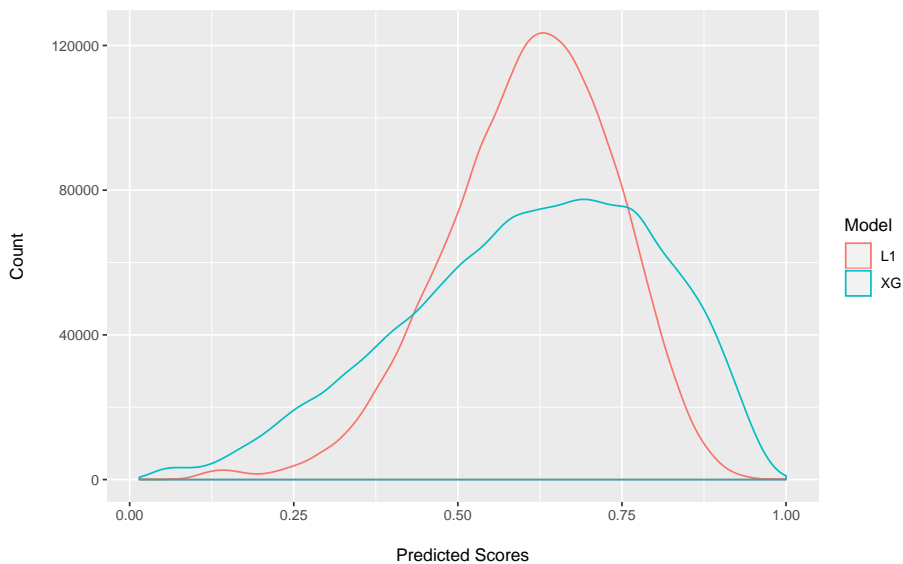}
\caption{\textit{Score Densities.} The figure compares the distributions of the scores predicted by two scoring models: L1-regularized logistic regression (red) and extreme gradient boosting (blue). Both models use the same data.}
\end{figure}

Logistic regression is a parametric learner which assumes a Gaussian distribution of the data. Because of this assumption, predicted probabilities can be output directly by the sigmoid function. In contrast, XG is a non-parametric learner which has more degrees of freedom and a higher potential for inductive bias reduction. Predicted scores produced by XG are not well calibrated \cite{niculescu2005obtaining}. Consider the example score distributions of L1 and extreme gradient boosting (XG) depicted in Figure 1. Here, adding regularization to logistic regression is important as we are dealing with high-dimensional data with noisy features. Compared to L1, the range of the output probabilities of XG is wider. 

We require the labeling model to produce well-calibrated probabilities as we limit the number of selected rejects based on the predicted PD values. Furthermore, by using different base models for application scoring and reject inference, we strive to reduce bias and error propagation. Hence, using a weak learner for reject inference is more promising. 

An important aspect of our framework is to account for a higher default rate among the rejects \cite{maldonado2010semi}. Recall that $X$ is partitioned into accepts and rejects based on a scoring model that is currently employed by a financial institution. Assuming that the scoring model in place performs better than a random loan allocation, we expect that the default rate among rejects is higher than among accepts. To address that difference, we introduce the imbalance parameter $\theta$ into our self-learning framework. On each labeling iteration, we only select the top $\alpha\%$ of the \textit{good} loans and top $\alpha \theta\%$ of the \textit{bad} loans among rejects for labeling. Keeping only the top-ranked instances ensures that we append rejects with high confidence in the assigned labels, reducing the potential amount of noise. By setting $\theta > 1$ we append more \textit{bad} cases to the training data, accounting for the imbalance. Parameter $\theta$ can be optimized at the parameter tuning stage.

Different variants of self-learning consider different ways to choose the most confident cases for labeling: either selecting top and bottom percentiles of the probability distribution or selecting cases based on a pre-defined probability threshold \cite{chapelle2006semi}. We suggest using the combined approach: on the first iteration, we compute the corresponding score values $c_{g}$ and $c_{b}$ for the selected $\alpha\%$ and $\alpha \theta\%$ probability percentiles. Since the labeling model is geared toward providing well-calibrated probabilities, we fix the absolute values $c_{g}$ and $c_{b}$ as thresholds for the subsequent iterations. By doing that, we reduce the risk of error propagation on further iterations. The absence of rejected cases with predicted scores above the fixed thresholds serves as a stopping criterion.

\begin{table}[H]\centering
	\begin{tabular*}{\textwidth}{llll}
		\hline
		\multicolumn{4}{l}{\textbf{Algorithm 2.} Shallow Self-Learning for Reject Inference} \\
		\hline
		1. & \multicolumn{3}{l}{Filter rejected cases in $X^r$ with isolation forest (see Algorithm 1).} \\
		2. & \multicolumn{3}{l}{Set $X^* = X^r$; set thresholds $c_b = \{\}$ and $c_g = \{\}$.} \\
		\multicolumn{4}{l}{\textbf{while} $X^* \ne \emptyset $ \textbf{do:}} \\
		& 3. & \multicolumn{2}{l}{Train L1 classifier $f(x)$ with penalty parameter $\lambda$ using all data in $X^a$.} \\
		& 4. & \multicolumn{2}{l}{Use $f(x)$ to predict PD for all unlabeled examples in $X^*$.} \\
		& \multicolumn{3}{l}{\textbf{if} $c_b = \{\}$ and $c_g = \{\}$ \textbf{do:}} \\
		&    & 5. & Derive $c_g$ such that $P(f(x^*_i \in X^*) < c_g) = \alpha$, $\alpha$ is a pre-defined percentile threshold. \\
		&    & 6. & Derive $c_b$ such that $P(f(x^*_i \in X^*) > c_b) = \alpha \theta$, $\theta$ is the imbalance parameter. \\
		& \multicolumn{3}{l}{\textbf{end if}} \\
		& 7. & \multicolumn{2}{l}{Select a subset of examples $X^* \subset X^r$ such that $f(x^*_i \in X^*) < c_g$ or $f(x^*_i \in X^*) > c_b$} \\
		& 8. & \multicolumn{2}{l}{Remove examples in $X^*$ from $X^r$.}\\
		& 9. & \multicolumn{2}{l}{Append examples in $X^*$ to $X^a$.} \\
		\multicolumn{4}{l}{\textbf{end while}} \\
		\hline
	\end{tabular*}
\end{table}

\subsection{Proposed Evaluation Measure}

Performance evaluation is an important part of selecting a suitable reject inference technique. In practice, accurate evaluation of reject inference is challenging. The true labels of rejected cases are unknown, which prohibits estimating the accuracy of a reject inference technique. Therefore, prior research evaluates the performance of a given technique by comparing the performance of the scoring model before and after appending the labeled rejected cases to the training data \cite{banasik2005credit,banasik2003sample,li2017reject}. 
The major downside of this approach is that the performance of a scoring model is not evaluated on a representative sample, which should include both accepts and rejects. Since labels of rejects are unknown, the literature suggests to evaluate models on a holdout sample that is drawn from the accepted cases (e.g., \cite{maldonado2010semi}). Very few empirical studies have access to the data on both accepts and rejects to evaluate scoring models \cite{crook2004does}.

Model selection based on the performance on accepts might lead to selecting a sub-optimal model. Let us illustrate that by comparing the performance of different scoring models validated on the accepts (4-fold cross-validation) and on the unbiased sample consisting of both accepts and rejects. We train a set of scoring models with different meta-parameter values and evaluate their performance in terms of the area under the receiver operating characteristic curve (AUC) \cite{rosset2004model}. Here, XG is used as a base classifier. Figure 2 depicts the results. 

The rank correlation between AUC values is just 0.0132. Due to the distribution differences between the accepted and rejected cases, the model's performance on the accepted applicants becomes a poor criterion for model selection. This result suggests that there is a need to develop an alternative measure for comparing and evaluating the scoring models in the presence of sample bias.

Without access to an unbiased sample that contains data on a representative pool of applicants, the literature suggests performing the evaluation by using synthetic data \cite{joanes1993reject}, emulating rejected cases by artificially moving the acceptance threshold \cite{maldonado2010semi} or using other criteria based on the applicants' feature values \cite{chen2001economic}. In this paper, we suggest the \textit{kickout} measure -- a novel evaluation measure that is based on the known data. We argue that developing such a measure is a valuable contribution since obtaining an unbiased data sample for performance evaluation is costly.

The key idea of the \textit{kickout} metric is to compare the set of applications accepted by a scoring model before and after reject inference. Recall that we have data on the previously accepted $X^a$ and rejected applicants $X^r$. Here, we partition $X^a$ into two subsets: $X^a_{train}$ and $X^a_{holdout}$. Let $C_1$ be a scoring model trained on $X^a_{train}$. We use $C_1$ to score cases from $X^a_{holdout}$ and select a pool of customers $A_1 \subset X^a_{holdout}$ that would be accepted by the model using the acceptance rate $\mu$. Thus, $A_1$ contains the (simulated) accepted applications before reject inference.

The rejected cases in $X^r$ are also split into two subsets: $X^r_{train}$ and $X^r_{holdout}$. The former is labeled with a given reject inference technique and appended to the $X^a_{train}$. Rejected cases in $X^r_{holdout}$ are appended to $X^a_{holdout}$, which now contains labeled accepts and unlabeled rejects, simulating the production-stage environment. Next, we train a new scoring model $C_2$ on the expanded training sample $X^a_{train}$ and use it to score and select customers in $X^a_{holdout}$ using the same acceptance rate $\mu$. Since both training and holdout samples have changed, model $C_2$ would accept a different pool of customers $A_2$. Analyzing the differences between $A_1$ and $A_2$, we can identify the kicked-out cases -- applications that were included in $A_1$ but do not appear in $A_2$.

\begin{figure}\centering
\label{cv_baseline}
\includegraphics[width = 0.55\textwidth, trim={0 0 0 0}, clip]{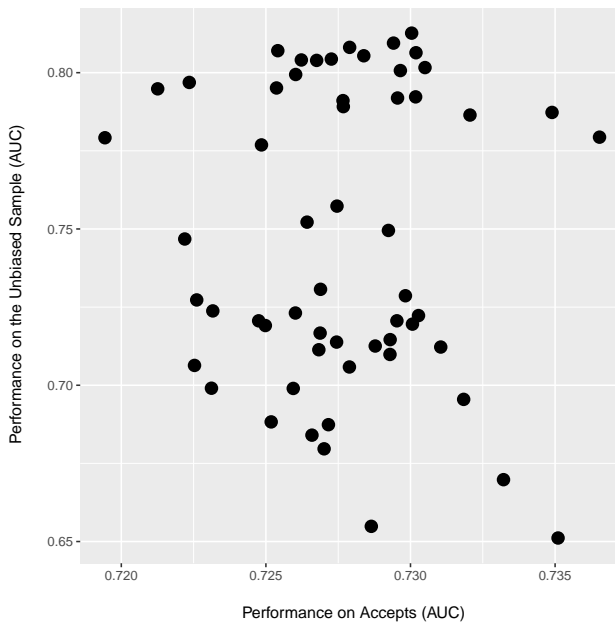}
\caption{Comparing AUC on the accepted cases (4-fold cross-validation) and the unbiased sample. The dots indicate scoring models with different meta-parameters.}
\end{figure}

We define the \textit{kickout} as follows:

\begin{equation}
\textnormal{\textit{kickout}} = \frac{\frac{\textnormal{K}_B}{\textnormal{p}(B)} - \frac{\textnormal{K}_G}{1 - \textnormal{p}(B)}}{\frac{\textnormal{S}_B}{\textnormal{p}(B)}},
\end{equation}
\vspace{3ex}

where $K_{B}$ is the number of \textit{bad} cases that were kicked out from the set of accepted cases after performing reject inference, $K_{G}$ is the number of kicked-out \textit{good} cases, $S_{B}$ is the number of \textit{bad} cases selected by the original model, and $\textnormal{P}(B)$ is the share of \textit{bad} cases in $A_1$. 

The \textit{kickout} metric ranges from $-1$ (all \textit{good} cases and no \textit{bad} cases are kicked out) to $1$ (all \textit{bad} cases and no \textit{good} cases are kicked out). We normalize the metric by the share of \textit{bad} cases to reflect the difficulty of kicking out a \textit{bad} customer. Positive values of \textit{kickout} signal a positive impact of reject inference, with higher values indicating a better performance of a given technique.

It is important to note that the \textit{kickout} metric does not require knowing the actual labels of the rejected cases that replace previously accepted cases. Instead, the metric focuses on the kicked-out applications. Replacing a \textit{bad} loan with a rejected case may have two possible outcomes. If the newly selected rejected case is also \textit{bad}, we are indifferent between the old and the new scoring model. If the rejected case is \textit{good}, the scoring model improves. Therefore, kicking out a \textit{bad} case has a positive expected value. In contrast, kicking out a \textit{good} case has a negative expected value: we are indifferent between the old and the new scoring model if the new rejected case is \textit{good}, whereas scorecard performance deteriorates if the rejected case is \textit{bad}. Hence, a good reject inference technique should change a scorecard such that it starts to kick out more \textit{bad} and less \textit{good} customers. 

The proposed measure relies on two assumptions. First, we assume that all \textit{bad} loans and all \textit{good} loans have the same expected value: that is, replacing one \textit{bad} case with another \textit{bad} case does not have any effect on the model's performance. Given the stable interest rates that determine the return on investment at fixed terms \cite{verbraken2014development} and an uncertain relationship between a loan amount and its PD, we argue that this assumption is reasonable in a credit scoring context. Second, we assume that the \textit{bad} ratio among rejected cases is higher compared to the accepted applications. As we detailed above, this assumption holds if the employed scoring model performs better than random.

\section{Experimental Results}

\subsection{Data Description}

The empirical experiments are based on a real-world credit scoring data set on consumer micro-loans provided by Kreditech, a Germany-based financial institution. The data set contains 2,410 features describing the applicants, their behavior and loan characteristics. The target variable is a binary indicator of whether the customer has repaid the loan. The data consist of 59,593 loan applications, out of which 39,579 were accepted and 18,047 were rejected. The target variable is only observed for the accepts, whereas the repayment status of rejects is unknown. Table \ref{data_stats} summarizes the main characteristics of the data set. 

In addition, we have access to a unique unbiased sample with 1,967 observations. Customers in that sample have been granted a loan without scoring. The sample, therefore, includes cases that would normally be rejected by a scorecard. This makes it representative of the through-the-door population of customers who apply for a loan. As noted in Table \ref{data_stats}, the default rate in the unbiased sample is 1.7 times higher than on the accepted cases. The unbiased sample allows us to evaluate the performance gains from reject inference on the sample that is representative of the production environment.

\subsection{Experimental Setup}

To evaluate the effectiveness of our propositions, we perform two experiments. Experiment 1 benchmarks the proposed self-learning framework against conventional reject inference techniques and standard self-learning. In the second experiment, we illustrate the effectiveness of the new \textit{kickout} measure for model selection. Below, we describe the modeling pipeline for these experiments. 

We partition the data into three subsets: accepts, rejects and the unbiased holdout sample consisting of both accepts and rejects. Next, we use 4-fold stratified cross-validation on the accepts to perform reject inference. On each iteration, the training folds are used to develop a reject inference technique that is used to infer labels of the rejects. Next, labeled rejects are appended to the training folds, providing a new sample to train a scoring model. Finally, a scoring model after reject inference is evaluated on the remaining fold and on the holdout sample. To ensure robustness, we evaluate performance on 50 bootstraped samples of the holdout set. Performance metrics of the reject inference techniques are then averaged over $4 \times 50$ obtained values.

We use XG classifier as a scoring model in both experiments. Meta-parameters of XG are tuned once on a small subset of training data using grid search. Within the experiments, we employ early stopping with 100 rounds while setting the maximum number of trees to 10,000 to fine-tune the model for each fold.

\vspace{2ex}
\begin{table}[H]\centering
	\caption{Data Summary}
	\label{data_stats}
	\vspace{-1ex}
	\begin{tabular}{@{\extracolsep{10pt}} llll}
		\hline
		\multicolumn{1}{c}{Characteristic} & \multicolumn{1}{c}{Accepts} & \multicolumn{1}{c}{Rejects} & \multicolumn{1}{c}{Unbiased} \\
		\hline
		Number of cases    & 39,579 & 18,047 & 1,967 \\
		Number of features & 2,410  & 2,410  & 2,410 \\
		Default rate       & 0.39   & unknown     & 0.66 \\
		\hline
	\end{tabular}
\end{table}

\begin{table}[H]\centering
	\caption{Reject Inference Techniques: Parameter Grid}
	\label{ri_grid} 
	\vspace{-1ex}
	\begin{tabular}{@{\extracolsep{15pt}} lll}
		\hline
		\multicolumn{1}{c}{Method} & \multicolumn{1}{c}{Parameter} & \multicolumn{1}{c}{Candidate values} \\
		\hline 
		\multirow{1}{*}{Ignoring rejects} & $-$ & $-$ \\
		\hline
		\multirow{1}{*}{Label all as \textit{bad}} & $-$ & $-$ \\
		\hline
		\multirow{1}{*}{Hard cutoff augmentation}  & threshold    & 0.3, 0.4, 0.5 \\
		\hline
		\multirow{2}{*}{Parcelling}  & multiplier    & 1, 2, 3 \\
		& no. batches       & 10 \\
		\hline
		\multirow{2}{*}{CV-based voting}  & threshold    & 0.3 \\
		& no. folds     & 2, 5, 10 \\
		\hline
		\multirow{2}{*}{Regular self-learning}  & max no. iterations & 5 \\
		& percentage     &  0.01, 0.02, 0.03\\
		\hline
		\multirow{4}{*}{Suggested self-learning} & filtered percentage & 0\%, 2\%  \\
		& max no. iterations & 5 \\
		& percentage     & 0.01, 0.02, 0.03 \\
		& multiplier     & 1, 2 \\
		\hline
	\end{tabular}
\end{table}

In Experiment I, we compare the suggested self-learning framework to the following benchmarks: ignore rejects, label all rejects as \textit{bad} risks, hard cutoff augmentation, parceling, cross-validation-based voting and standard self-learning. Here, cross-validation-based voting is an adaption of a label noise correction method suggested by \cite{verbaeten2003ensemble}. It refers to an extension of hard cutoff augmentation that employs a homogeneous ensemble of classifiers based on different training folds instead of a single scoring model to label the rejects. The labels are only assigned to the cases for which all individual models agree on the label. 

We test multiple versions of each reject inference technique with different meta-parameter values using grid search. For shallow self-learning, penalty $\lambda$ of the labeling model is tuned and optimized once on the first labeling iteration. Table \ref{ri_grid} provides the candidate values of meta-parameters.

For performance evaluation, we use three metrics that capture different dimensions of the predictive performance: AUC, Brier Score (BS) and R-Precision (RP). We use AUC as a well-known indicator of the discriminating ability of a scoring model. In contrast, BS measures the calibration of the predicted default probabilities. Last, we use RP as it better reflects the business context. The financial institution that provided data for this study decides on a loan allocation by approving a certain percentage of the least risky customers. RP measures performance only for cases which will indeed be accepted. In our experiments, we compute RP in the top $30\%$ of the applications with the lowest predicted default probabilities.

In Experiment II, we compare different variants of self-learning using grid search within the cross-validation framework described above. Apart from the three selected performance measures, we also evaluate reject inference in terms of the proposed \textit{kickout} measure. The goal of this experiment is to compare model rankings based on three evaluation strategies: performance on the accepted cases, performance on the unbiased holdout sample and performance in terms of the \textit{kickout} measure.

\subsection{Empirical Results}

\subsubsection{Experiment I: Assessing the Shallow Self-Learning}
 
\ \newline
Table \ref{ri_perf} summarizes the performance of the considered reject inference techniques. We compare the performance values on the accepted cases (using 4-fold cross-validation) and on the unbiased sample. Recall that the latter serves as a proxy for the production-stage environment for a scoring model, whereas the cross-validation performance refers to a conventional approach towards evaluation in credit scoring. 

According to the results, not all methods improve on the benchmark of ignoring rejects: only three out of six techniques achieve higher AUC and lower BS on the unbiased sample, and only one (shallow self-learning) has a higher RP. Labeling rejects as \textit{bad} risks performs better than disregarding reject inference on the accepts but does worse on the unbiased sample. In contrast, parceling is outperformed by all the remaining techniques on the accepts, whereas it has a higher AUC on the unbiased sample. These results support the argument that the scorecard performance on the accepted cases might be a poor indicator of the production-stage performance. 

Regular self-learning outperforms ignoring rejects in terms of AUC and BS but does not improve in terms of RP. The proposed shallow self-learning framework performs best in all three measures on the unbiased sample as well as on the accepted applicants. The best performance is achieved by a self-learning model that includes filtering. Therefore, we can conclude that the suggested modifications help to adjust self-learning for the reject inference problem.

\begin{table}\centering
    \caption{Comparing Performance of Reject Inference Techniques}
    \label{ri_perf} 
    \vspace{-1ex}
    \begin{tabular}{@{\extracolsep{5pt}} lcccccc}
        \hline
        \multirow{2}{*}{Method} & \multicolumn{3}{c}{Accepted cases} & \multicolumn{3}{c}{Unbiased sample} \\
         & AUC & BS &  RP & AUC & BS & RP \\
        \hline 
        Ignore rejects                    & 0.7297 & 0.1829 & 0.8436 & 0.8007 & 0.2092 & 0.7936\\
        Label all as bad                  & 0.7332 & 0.1816 & 0.8474 & 0.6797 & 0.2284 & 0.7253\\
        Hard cutoff augmentation          & 0.7295 & 0.1770 & 0.8430 & 0.7994 & 0.2212 & 0.7751\\
        Parcelling                        & 0.7277 & 0.1842 & 0.8430 & 0.8041 & 0.1941 & 0.7851\\
        CV-based voting                   & 0.7293 & 0.1804 & 0.8430 & 0.7167 & 0.2160 & 0.7510\\
        Regular self-learning             & 0.7302 & 0.1758 & 0.8434 & 0.8063 & 0.1838 & 0.7929\\
        \textbf{Shallow self-learning}  & \textbf{0.7362} & \textbf{0.1736} & \textbf{0.8492} & \textbf{0.8070} & \textbf{0.1799} & \textbf{0.7996} 
        \\
        \hline
    \end{tabular}
\end{table}
\vspace{2ex}

Performance gains appear to be modest, supporting the prior findings \cite{hand1993can}. We check statistical significance of the differences using Friedman's rank sum test and Nemenyi pairwise test \cite{garcia2010advanced}. According to Friedman's test, we reject the null hypothesis that all reject inference techniques perform the same at 5\% level for AUC ($\chi^2$ = 419.82), RP ($\chi^2$ = 326.99) and BS ($\chi^2$ = 485.59). Nemenyi test indicates that shallow self-learning performs significantly better than all competitors in terms of AUC and RP, whereas differences in BS between standard and shallow self-learning are not statistically significant at 5\% level.

Even small differences might have a considerable effect on the costs of the financial institution \cite{schebesch2008using}. Comparing shallow self-learning to ignoring rejects, 0.006 increase in RP translates to 60 less defaulted loans for every 10,000 accepted clients. Considering the average personal loan size of $\$17,100$ and interest rate of 10.36\% observed in the US in Q1 2019\footnote{Source: https://www.supermoney.com/studies/personal-loans-industry-study/}, potential gains from reject inference could amount for up to $\$1.13$ million depending on the recovery rates.

\subsubsection{Experiment II: Evaluation Strategy for Model Selection}
\ \newline
\noindent
In the second experiment, we perform model selection on 28 variants of self-learning with different meta-parameter values. Table \ref{es_cor} displays the correlation between model ranks in terms of three evaluation measures: AUC on the accepts, AUC on the unbiased sample and the \textit{kickout} measure.

According to Table \ref{es_cor}, the absolute value of rank correlations between the performance on the accepts and performance on the unbiased data does not exceed $0.01$. In contrast, the rankings based on \textit{kickout} are positively correlated with those on the unbiased sample ($r = 0.41$). Therefore, the common practice to assess reject inference strategies using the model's performance on the accepted cases provides misleading results as there is a very small correlation between the performance on the accepts and the performance on the production stage. In contrast, comparing reject inference techniques using the proposed \textit{kickout} measure is more promising.

Figure 3 illustrates the advantages of using \textit{kickout} instead of the performance on the accepts for model selection. Red points indicate the predictive performance of a scoring model selected by the \textit{kickout} measure, while green dots refer to the best-performing model on the accepts in terms of AUC, BS and RP. As before, we evaluate the selected scoring models on the unbiased sample.

\vspace{0ex}
\begin{table}[H]\centering
	\caption{Correlation between Evaluation Strategies}
	\label{es_cor} 
	\vspace{-1ex}
	\begin{tabular}{@{\extracolsep{15pt}} lccc}
		\hline
		\multicolumn{1}{c}{Evaluation strategy} & (1) & (2) & (3) \\
		\hline 
		(1) AUC on the accepted cases      & $1$       &              &      \\
		(2) AUC on the unbiased sample     & $-0.0009$    & $1$       &      \\
		(3) The kickout metric             & $0.0336$    & $0.4069$ &  $1$   \\
		\hline
	\end{tabular}
	\vspace{-2ex}
\end{table}
\vspace{2ex}

As shown in Figure 3, using the \textit{kickout} measure results in selecting a better model in terms of all three performance indicators. By relying on \textit{kickout} instead of the performance on the accepts, we are able to identify a scorecard that has a better performance on the unbiased sample.

\begin{figure}
\begin{center}
\label{sl_selection}

\includegraphics[width = 0.496\textwidth, trim={0 0 0 0}, clip]{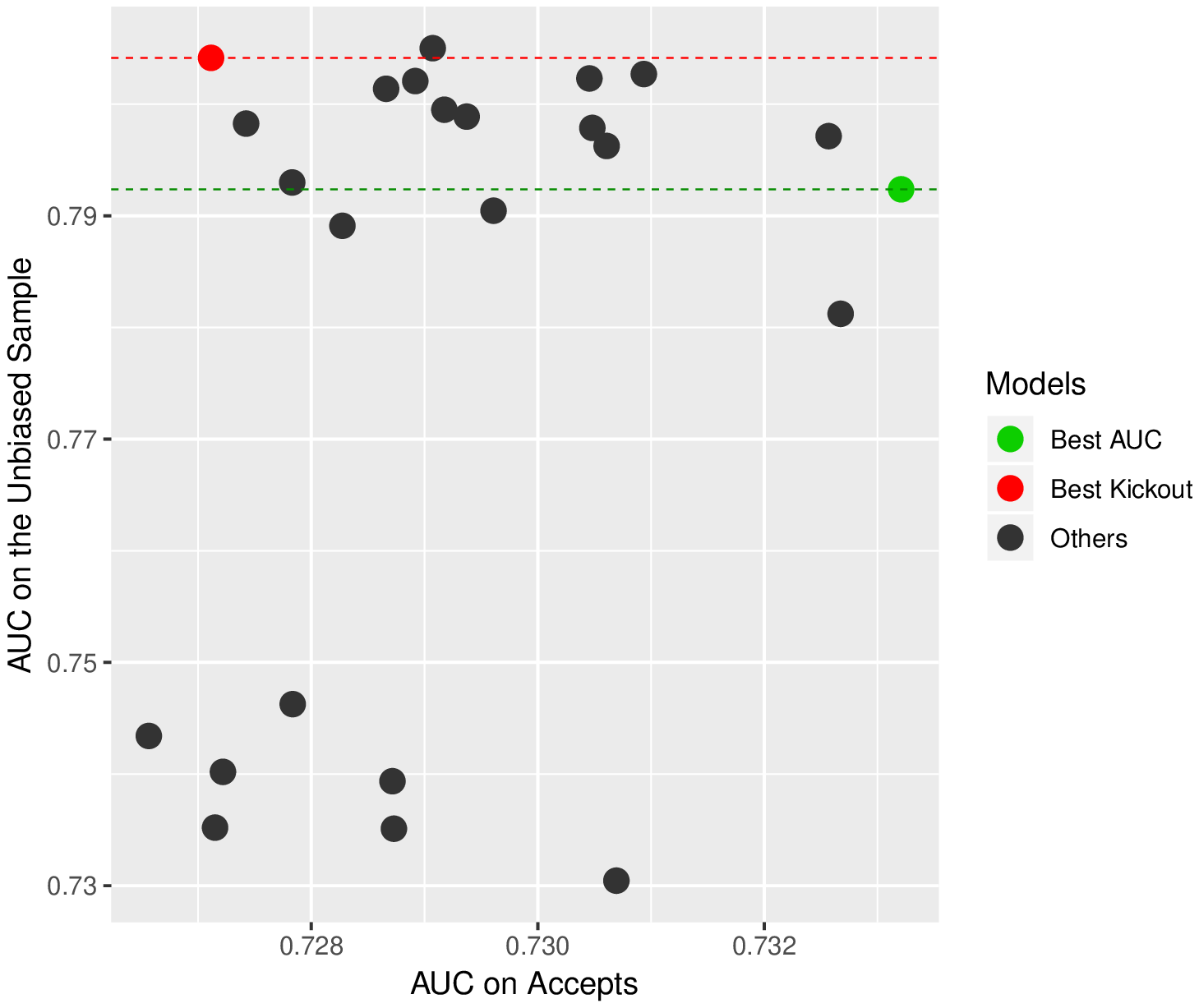}
\includegraphics[width = 0.496\textwidth, trim={0 0 0 0}, clip]{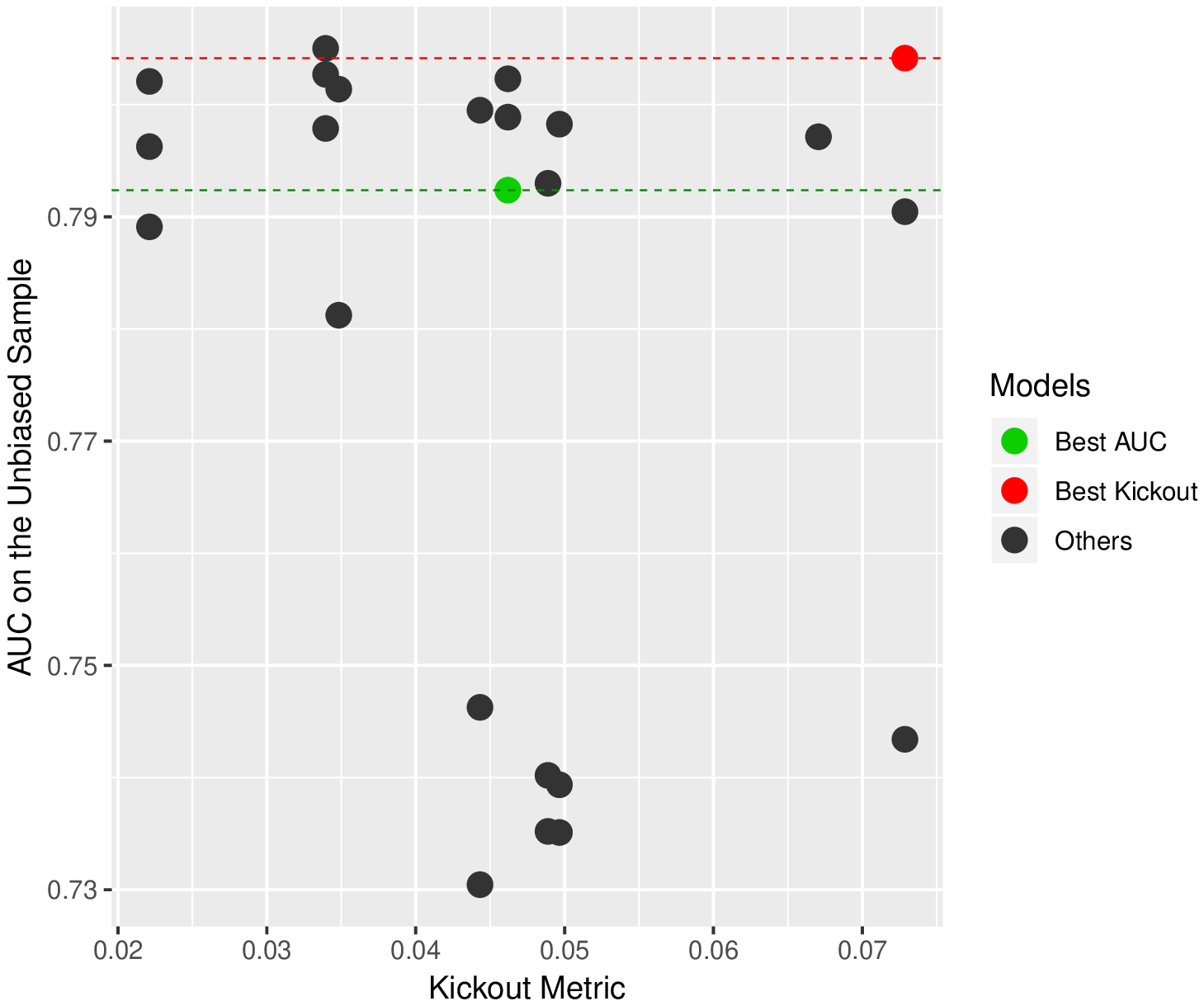}

\vspace{3ex}

\includegraphics[width = 0.496\textwidth, trim={0 0 0 0}, clip]{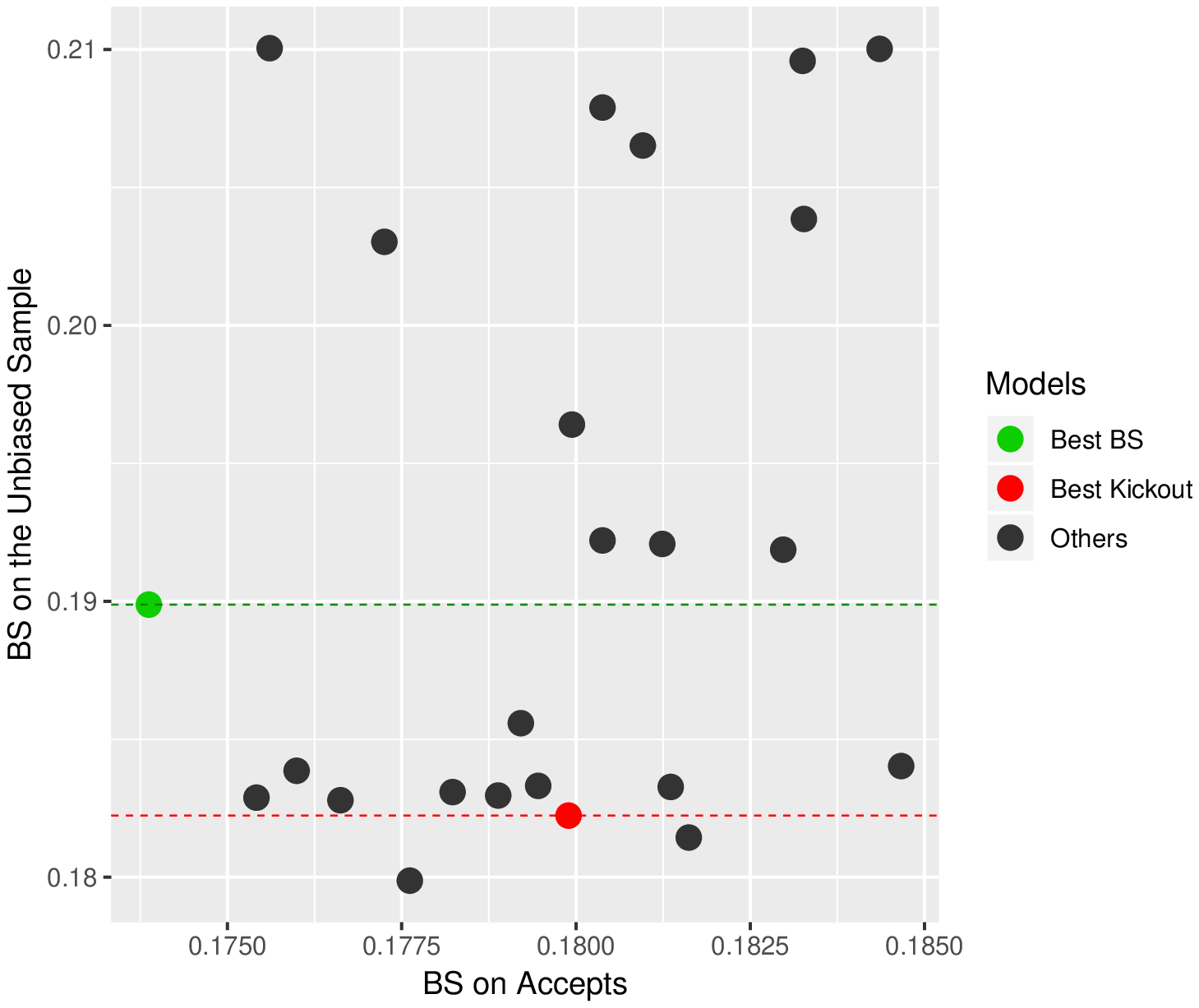}
\includegraphics[width = 0.496\textwidth, trim={0 0 0 0}, clip]{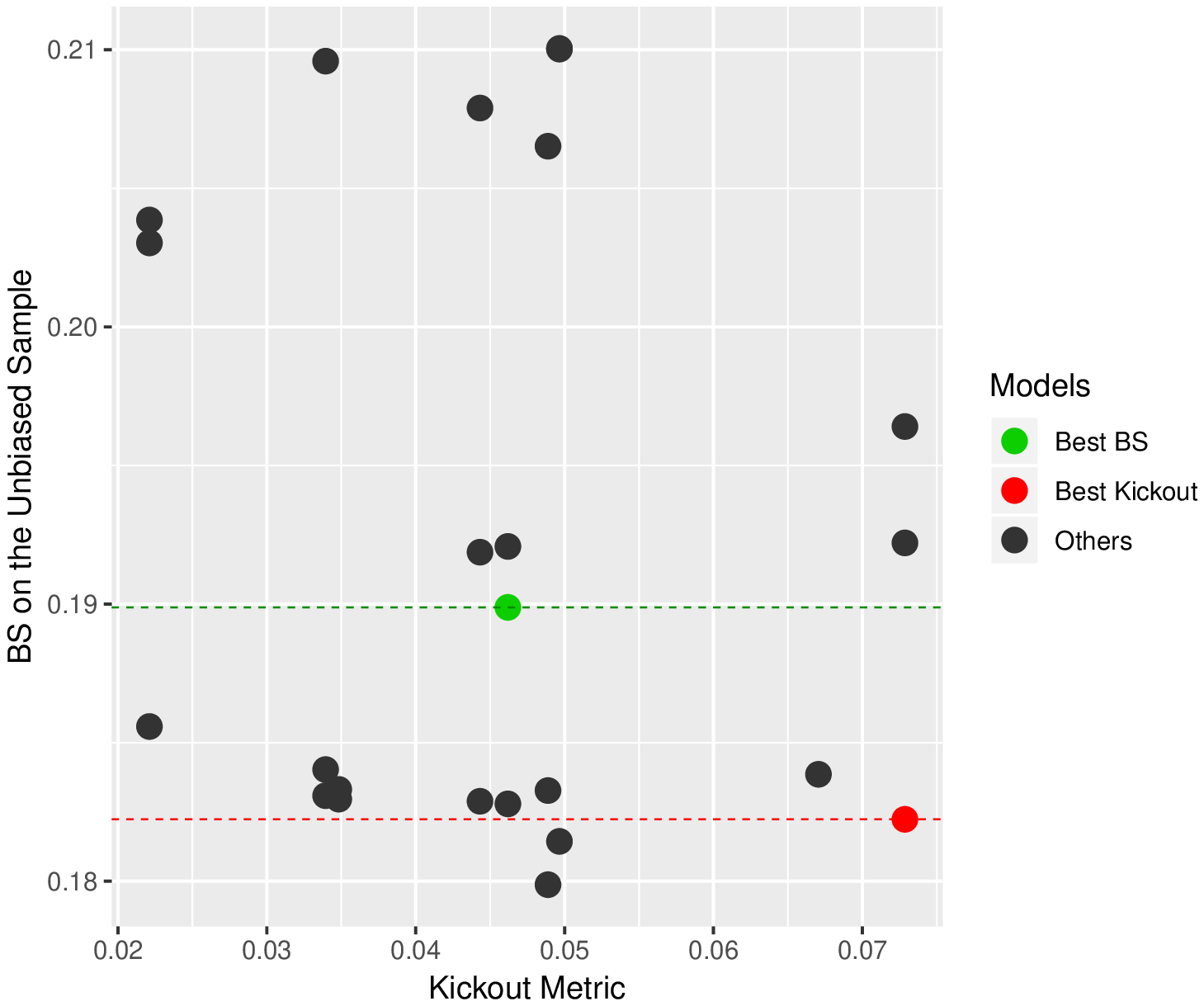}

\vspace{3ex}

\includegraphics[width = 0.496\textwidth, trim={0 0 0 0}, clip]{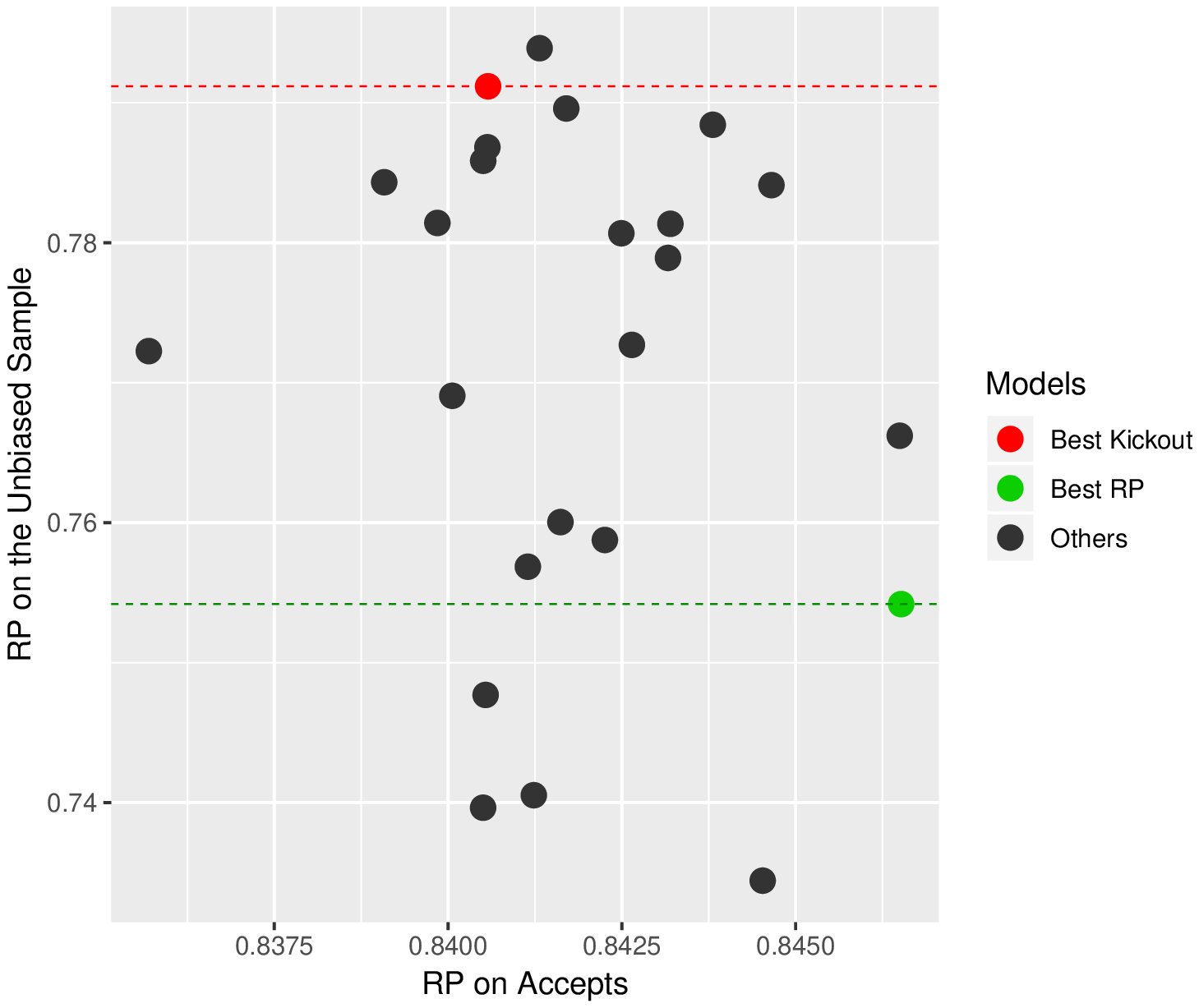}
\includegraphics[width = 0.496\textwidth, trim={0 0 0 0}, clip]{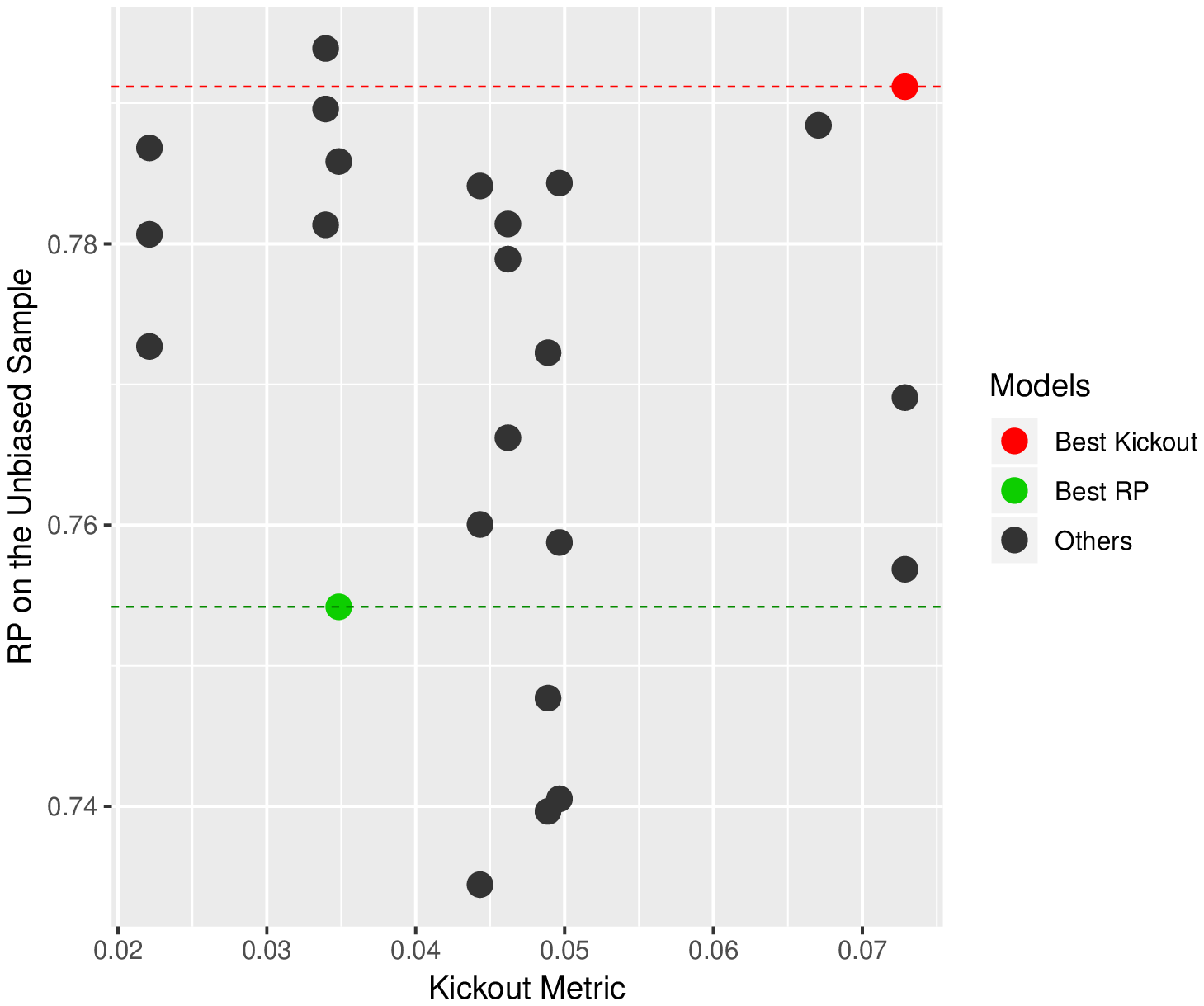}

\vspace{-1ex}

\caption{\textit{Model Selection Results.} Two upper diagrams compare model selection results based on AUC on the accepted cases (green) and on the kickout metric (red). Two diagrams in the center compare results based on Brier Score on the accepts (green) and the kickout metric (red), whereas two lower diagrams refer to R-Precision on the accepts (green) and kickout (red). }
\end{center}
\end{figure}

These results emphasize the importance of using a suitable evaluation strategy to assess the value of reject inference. Relying on conventional evaluation measures such as AUC that are estimated on the accepted cases would result in selecting a suboptimal scoring model in terms of its production-stage performance. Our experiments show that \textit{kickout} proves to be a suitable measure for doing model selection. According to the results, the \textit{kickout} measure identifies a better scoring model in the absence of an unbiased sample, which is particularly useful for practitioners.

\section{Conclusion}

This paper suggests a self-learning framework with distinct training and labeling regimes for reject inference in credit scoring and develops a novel evaluation measure for model selection. We evaluate the effectiveness of our approach by running empirical experiments on a high-dimensional real-world credit scoring data set with unique properties.

Empirical results indicate that the proposed self-learning framework outperform regular self-learning and conventional reject inference techniques in terms of three performance measures. These results indicate that the modifications suggested here help to adjust self-learning to the reject inference problem. 

We also develop a novel evaluation measure to perform model selection for reject inference techniques. We show that the standard practice of selecting models (or meta-parameters) based on their performance on the accepted cases may lead to choosing a model with a suboptimal predictive performance at the production stage. Compared to the standard approach, the proposed \textit{kickout} measure exhibits a higher correlation with the performance on the unbiased sample and allows to identify a scoring model with better performance.

Our results imply that future research on reject inference should not rely on the model's performance on the accepted cases to judge the value of a certain reject inference technique. The \textit{kickout} measure proves to be a good alternative for practitioners who often do not have access to an unbiased sample that contains both accepted and rejected applications.




\bibliographystyle{unsrt}

\end{document}